\documentclass[conference]{IEEEtran}
\IEEEoverridecommandlockouts
\usepackage{cite}
\usepackage{amsmath,amssymb,amsfonts}
\usepackage{algorithmic}
\usepackage{graphicx}
\usepackage{textcomp}
\usepackage{xcolor}
\usepackage{multirow}
\usepackage{url}

\def\BibTeX{{\rm B\kern-.05em{\sc i\kern-.025em b}\kern-.08em
    T\kern-.1667em\lower.7ex\hbox{E}\kern-.125emX}}
\begin{document}

\title{Faces of Experimental Pain: Transferability of Deep Learned Heat Pain Features to Electrical Pain *\\
\thanks{Identify applicable funding agency here. If none, delete this.}
}

\author{\IEEEauthorblockN{Pooja Prajod}
\IEEEauthorblockA{\textit{Lab for Human-Centered AI} \\
\textit{University of Augsburg}\\
Augsburg, Germany \\
pooja.prajod@uni-a.de}
\and
\IEEEauthorblockN{Dominik Schiller}
\IEEEauthorblockA{\textit{Lab for Human-Centered AI} \\
\textit{University of Augsburg}\\
Augsburg, Germany \\
dominik.schiller@uni-a.de}
\and
\IEEEauthorblockN{Daksitha Withanage Don}
\IEEEauthorblockA{\textit{Lab for Human-Centered AI} \\
\textit{University of Augsburg}\\
Augsburg, Germany \\
daksitha.withanage.don@uni-a.de}
\and
\IEEEauthorblockN{Elisabeth André}
\IEEEauthorblockA{\textit{Lab for Human-Centered AI} \\
\textit{University of Augsburg}\\
Augsburg, Germany \\
elisabeth.andre@uni-a.de}
}

\maketitle

\begin{abstract}
The limited size of pain datasets are a challenge in developing robust deep learning models for pain recognition. Transfer learning approaches are often employed in these scenarios. In this study, we investigate whether deep learned feature representation for one type of experimentally induced pain can be transferred to another. Participating in the AI4Pain challenge, our goal is to classify three levels of pain (No-Pain, Low-Pain, High-Pain). The challenge dataset contains data collected from 65 participants undergoing varying intensities of electrical pain. We utilize the video recording from the dataset to investigate the transferability of deep learned heat pain model to electrical pain. In our proposed approach, we leverage an existing heat pain convolutional neural network (CNN) - trained on BioVid dataset - as a feature extractor. The images from the challenge dataset are inputted to the pre-trained heat pain CNN to obtain feature vectors. These feature vectors are used to train two machine learning models: a simple feed-forward neural network and a long short-term memory (LSTM) network. Our approach was tested using the dataset's predefined training, validation, and testing splits. Our models outperformed the baseline of the challenge on both the validation and tests sets, highlighting the potential of models trained on other pain datasets for reliable feature extraction. 




\end{abstract}


\section{Introduction}

Pain assessment is a complex yet critical aspect of healthcare, influencing treatment strategies and patient well-being. Traditional methods of pain evaluation often rely on subjective patient reports, which can be inconsistent and difficult to quantify objectively~\cite{cowen2015assessing}. In the recent years, researchers have demonstrated the potential of machine learning methods in pain assessment by analyzing various behavioral and physiological data.

The International Association for the Study of Pain (IASP) describes pain as ``an unpleasant sensory and emotional experience associated with, or resembling that associated with, actual or potential tissue damage"~\cite{raja2020revised}. While many researchers have raised the need for refining this definition, most of them agree on the presence of an emotional component in pain~\cite{cohen2018reconsidering}. Similar to emotions, pain manifests through various behavioral (e.g., facial expressions, body pose) and physiological changes (e.g., elevated heart rate, breathing rate).

Among the various modalities, facial expressions of pain is particularly interesting for pain recognition models. Research has shown that pain expressions are perceived similarly between different cultures (e.g., western vs. eastern cultures)~\cite{klingner2023mimik}. Moreover, pain expressions involve certain muscles - especially around the eyes - that are not controlled voluntarily~\cite{williams2011FacialEO, hadjistavropoulos2011biopsychosocial}. This implies that pain facial expressions are difficult to be completely masked or faked, making them a reliable modality for recognizing pain. Furthermore, pain facial expressions can be analyzed using RGB images/videos, which represent a non-intrusive and non-contact solution.

Previous research has shown that deep learning models perform better than hand-crafted features (e.g., facial action units) in predicting pain~\cite{zamzmi2018neonatal, egede2017fusing}. However, large amounts of data are required to train robust deep learning models. So, training deep pain recognition models are challenging due to the typical small size of pain datasets~\cite{wang2017regularizing,  hassan2019automatic, kunz2017problems, prajod2022deep, xiang2022imbalanced}. To mitigate this, transfer learning approaches are often employed. These approaches involve utilizing features learned for a task in another related task. 

In this paper, we investigate the transferability of facial pain features for experimentally induced pain datasets. Specifically, we assess whether the features learned for binary pain detection in a heat pain dataset can be leveraged to train a model for electrical pain. Aligning with this goal, we participated in the AI4Pain challenge~\cite{challenge_paper}, which involved developing a model for three-class pain recognition. The challenge dataset comprises facial videos and functional near-infrared spectroscopy (fNIRS) data collected from 65 participants undergoing varying intensities of electrical pain stimuli. Since our focus was on transferability, we utilized only the facial videos as it is a popular modality used in other pain datasets.

Our approach involves utilizing an existing convolutional neural network (CNN) trained on a heat pain dataset as a feature extractor. The face-cropped images from the AI4Pain dataset are fed to this pre-trained model for producing feature vectors, which are used to train three-class (No-Pain, Low-Pain, High-Pain) pain recognition models. We validate our approach using training two AI4Pain models: a simple artificial neural network (ANN) with majority voting and a long short-term memory (LSTM) network. Both models outperformed the challenge baseline on validation and test sets. Our results indicate that features learned by an experimental pain model can be transferred for pain recognition in another experimental pain stimuli.

\section{Related Work}

Although the AI4Pain challenge dataset contains video and fNIRS recordings, we focus on pain recognition through facial images/video. So, we discuss some of the works that utilized facial expressions to predict pain, especially focusing on transfer learning. Since our work relies on pain features learned from another pain dataset, we also discuss the findings of previous works that investigated cross-dataset applicability of pain models. 

\subsection{Transfer Learning in Facial Pain Recognition}
Facial expressions is a behavioral signal that is the most informative form of nonverbal communication~\cite{valstar201711, ko2018brief}. The development of facial action coding system (FACS) led to a surge in identifying facial expression patterns of pain~\cite{prkachin2009assessing}. The recent advancements in machine learning methods led to an increasing focus on automatic pain recognition systems. 

Florea~et~al.~\cite{florea2014learning} presented one of the earlier works on transfer learning in pain prediction. They first trained a model to learn data distribution of hand-crafted geometric features for an emotion recognition task. These learned representations were utilized in training a support vector machine in pain intensity estimation.

The state-of-the-art pain facial expression recognition models rely on deep learning methods that learn relevant features from the raw input~\cite{gkikas2023automatic}. However, deep learning models are typically trained on large datasets, whereas pain datasets are often small. As mentioned above, transfer learning approaches are often employed to circumvent this limitation. One transfer learning technique is to use pre-trained models as feature extractors and subsequently training pain recognition models on the extracted features. For example, Zamzmi~et~al.~\cite{zamzmi2018neonatal} used a face recognition model (VGG-Face) and object recognition models (VGG trained on ImageNet dataset) as feature extractors. The features obtained from the CNN models were used to train a neonatal pain recognition model. Similarly, Egede~et~al.~\cite{egede2017fusing} utilized an action unit (AU) detection CNN as a feature extractor to estimate pain in a shoulder pain dataset. Both these works found that CNN-extracted features outperformed the hand-crafted features.

Another transfer learning technique involves fine-tuning the pre-trained CNN using the target dataset. Unlike the feature extractor method, this technique changes the learned representation of the source model. Examples of fine-tuning method include the works of Wang~et~al.~\cite{wang2017regularizing} and Prajod~et~al.~\cite{prajod2022deep, prajod2022using}. Wang~et~al. fine-tuned the VGG-Face CNN  for pain estimation, whereas Prajod~et~al. transfer learned pain detection from emotion recognition CNN. Their models showcased good performances, plausibly due to the similarity between the source and target tasks.

Most existing facial pain datasets consists of consecutive video frames or pain video snippets. These datasets contain temporal information that are seldom used by CNNs. Hence, some studies proposed LSTMs to capture time-series data from the pain videos/image sequences. For example, Rodriguez~et~al.~\cite{rodriguez2017deep} used fine-tuned the VGG-Face network using pain images and used the fine-tuned network as a feature extractor. The feature vectors from the CNN were then used to train an LSTM for a binary pain detection task. The LSTM architecture led to an improved performance compared to frame-wise CNN predictions. However, Haque~et~al.~\cite{haque2018deep} obtained a contradicting result. They fine-tuned the VGG-Face network using pain images and connected the CNN network to LSTM layers. They used RGB, depth information, and thermal video frames as input images. Their results showed that incorporating temporal information did not improve pain prediction. 

The works discussed above predominantly utilize VGG-Face or facial expression models (emotions, AU) for extracting features for pain recognition. In our work, we use an existing pain detection model as a feature extractor and subsequently train a pain recognition model on the challenge dataset.

\subsection{Cross-Dataset Pain Applicability}

Existing pain datasets can be broadly classified as clinical and experimental pain~\cite{kunz2019facial, prajod2022using}. Clinical pain originates from existing clinical conditions such as surgery or back pain, whereas experimental pain is induced through stimuli like heat or electricity. While selecting a pain detection model as a feature extractor, it is important to ensure that the model is capable of detecting features relevant for the given dataset. One indicator is good cross-dataset performances of a model, which suggests that the model learned generic pain features. 

There are limited number of works that perform cross-dataset evaluations of pain models. To address this limitation, Othman~et~al.~\cite{othman2019cross} conducted cross-dataset evaluations of their pain models trained on a heat pain dataset and an electrical pain dataset. Both heat and electrical pain models achieved comparable performances in both within-dataset and cross-dataset evaluations.

Dai~et~al.~\cite{dai2019real} trained pain detection models using a shoulder pain dataset. They tested their models on a heat pain dataset and found a decline in performances. Similarly, Tavakolian~et~al.~\cite{tavakolian2020self} investigated the same datasets and assessed cross-dataset performances of both models. They observed lower cross-dataset performance for both models. Furthermore, Prajod~et~al.~\cite{prajod2022using} also conducted cross-dataset evaluations on models trained on the same datasets. They observed that the heat pain dataset was more robust and achieved comparable within-dataset and cross-dataset evaluations. They further validated the features learned by the models using explainable AI techniques. 

The above results indicate that a clinical pain dataset (e.g., shoulder pain) may not be ideal for representing features of electrical pain images. On the other hand, the findings of Othman~et~al. suggest that two experimental pain models (e.g., heat and electrical pain) may be more compatible.

\section{Approach}

\begin{figure*}[htbp]
\centerline{\includegraphics[width=0.9\textwidth]{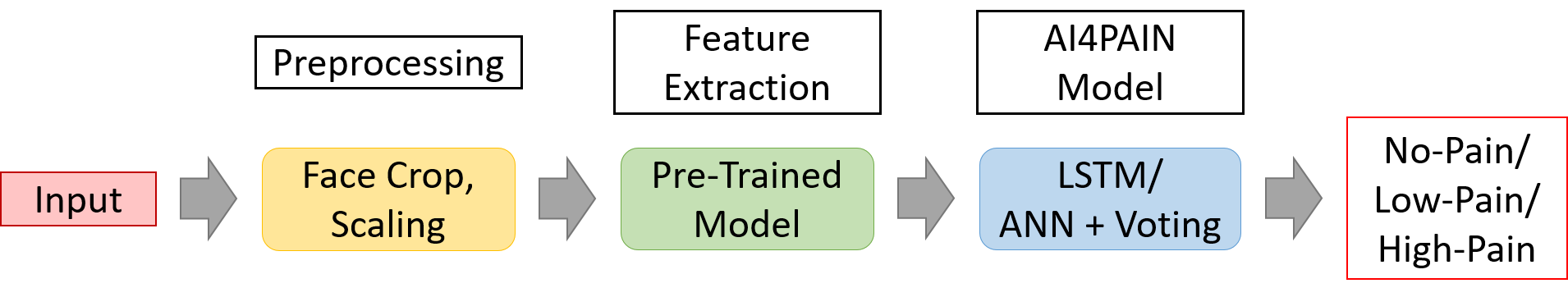}}
\caption{An illustration of the machine learning pipeline with an overview of the various components.}
\label{fig:pain_pipeline}
\end{figure*}

Our model development follows a typical machine learning pipeline involving pre-processing, feature extraction, and training phases. The pre-processing involves face cropping and scaling of the input images. The relevant facial pain features were extracted using a pre-trained CNN which was trained on another pain dataset. This pre-trained model was chosen because it has been demonstrated to learn well-known facial pain patterns. In the training phase, we trained two different pain recognition models to evaluate our approach. These steps followed in each phase are detailed in Sections~\ref{sec:facecrop}, \ref{sec:feats}, and \ref{sec:training}, respectively. Figure~\ref{fig:pain_pipeline} shows an overview of the various components involved in the model development.

\subsection{Dataset}

The AI4Pain dataset includes data from 65 participants and is designed to distinguish between three pain categories: No-Pain, Low-Pain, and High-Pain.

This dataset consists of recordings from 23 women and 42 men, with an average age of 29.06 years (SD = 8.28). Participants were seated with their arms resting on a table.
Each participant wore a functional near-infrared spectroscopy (fNIRS) headset positioned on the frontal area, and their facial video data was captured using a Logitech StreamCam at a sampling rate of 30Hz.

Transcutaneous electrical nerve stimulation (TENS) electrodes were placed on the inner forearm and the back of the hand of the participants' right arms. 
To prevent habituation and order effects, the location and intensity of the pain stimuli were counterbalanced. Each session began with approximately 60-second baseline period representing the No-Pain condition. 
The experimental design included six repetitions of each stimulus (Low Pain and High Pain) at each anatomical location (Arm and Hand), totaling 12 repetitions for both Low Pain and High Pain stimuli. 
Specific stimulus locations have been omitted from the dataset. The dataset collection procedure has been detailed in~\cite{fernandez2023multimodal}.
The dataset is divided into three sets \emph{Train} (41 participants), \emph{Validation} (12 participants), \emph{Test} (12 participants
For our experiments, we use the \emph{Train} for training and \emph{Validation} splits for hyper parameter optimization. 
Results are reported for \emph{Validation} and \emph{Test}.


\subsection{Preprocessing}
\label{sec:facecrop}
The preprocessing step involves extracting and cropping facial regions from video frames, and preparing the data for subsequent analysis. Initially, the face-alignment~\cite{face-alignment} library detects facial landmarks in each video frame, ensuring accurate identification of facial regions. Once the landmarks are detected, a bounding box is calculated around the face, which is then used to crop the face from the frame. The cropped faces are resized to a standard target size of $224 \times 224$ pixels to ensure uniform input dimensions for further processing. Videos are read frame-by-frame using OpenCV\footnote{\url{https://pypi.org/project/opencv-python/}}, and face detection and cropping procedures are applied for each frame. The videos are processed at a resampled frame rate of 30 frames per second (fps) to maintain uniformity across the dataset. Each cropped face is saved as an image in a specified output directory, resulting in a series of images representing the facial expressions over time.

\subsection{Feature Extraction}
\label{sec:feats}

Recent facial expression recognition systems employ deep learning models that are capable of extracting features from the raw images and subsequently making predictions. However, such models need to be trained on large amounts of data for reliable performance. But pain datasets are typically small due to ethical considerations and data collection challenges involved in inducing pain~\cite{wang2017regularizing, hassan2019automatic, kunz2017problems, xiang2022imbalanced}. In these cases, prior research~\cite{wang2017regularizing, prajod2022deep, coutrin2022convolutional} adopted a transfer learning approach to train deep learning models using small pain datasets. These studies leverage the weights of a CNN model trained on another dataset (source) for training the target model. In other words, the features learned for the source task are transferred to the target task.

A special case of this type of transfer learning is utilizing the source model as a feature extractor~\cite{florea2014learning, egede2017fusing, zamzmi2018neonatal, prajod2022deep}. The weights of the source models are intact and the output produced by the model for an input image is used as feature vectors for a target model. This approach is similar to traditional machine learning models that rely on hand-crafted features, though the features in this case are obtained through a pre-trained model.

The ideal feature extractor would be a model that generates features relevant to pain recognition. A straightforward choice is an existing pain model trained on another dataset that performs well in predicting pain. We use the features extracted from this existing pain model to train a simpler model that predicts pain in the target dataset. However, it is important to note that multiple images/videos of the pain datasets originate from the same participant, which reduces the variations in the pain responses. Hence, well-performing existing pain models may learn features specific to their training datasets. To mitigate this risk, the ideal feature extraction model should be validated on other datasets (cross-dataset evaluations). 

\begin{figure*}[htbp]
\centerline{\includegraphics[width=0.7\textwidth]{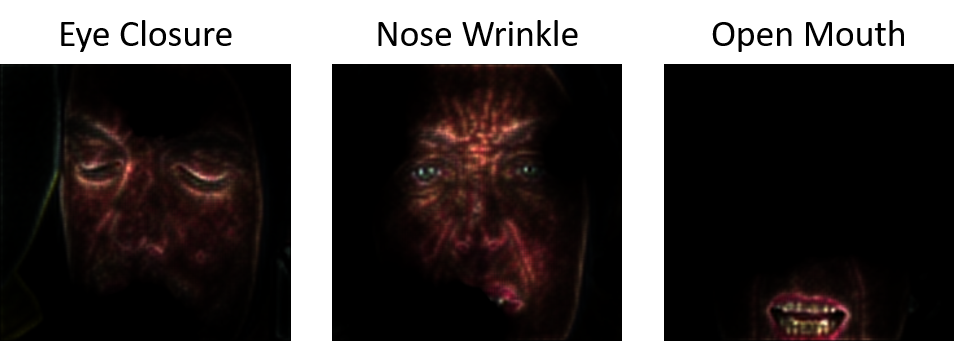}}
\caption{Few saliency maps taken from Prajod~et~al.~\cite{prajod2022using} showing the features learned by their model.}
\label{fig:pain_patterns}
\end{figure*}

We considered a few existing models that have shown good cross-dataset performances. In this paper, we leveraged an existing model developed by Prajod~et~al.~\cite{prajod2022using} for pain detection. This is because, in addition to cross-dataset evaluations, the authors also applied explainable AI methods to visualize the learned representations of their model. We note that the saliency maps they presented correspond to some of the established facial expression patterns. For example, their saliency maps (see Figure~\ref{fig:pain_patterns}) showed patterns typically associated with pain expressions such as nose wrinkling, lips apart, and eye closure~\cite{prkachin2009assessing, kunz2019facial}.

\begin{figure}[htbp]
\centerline{\includegraphics[width=\linewidth]{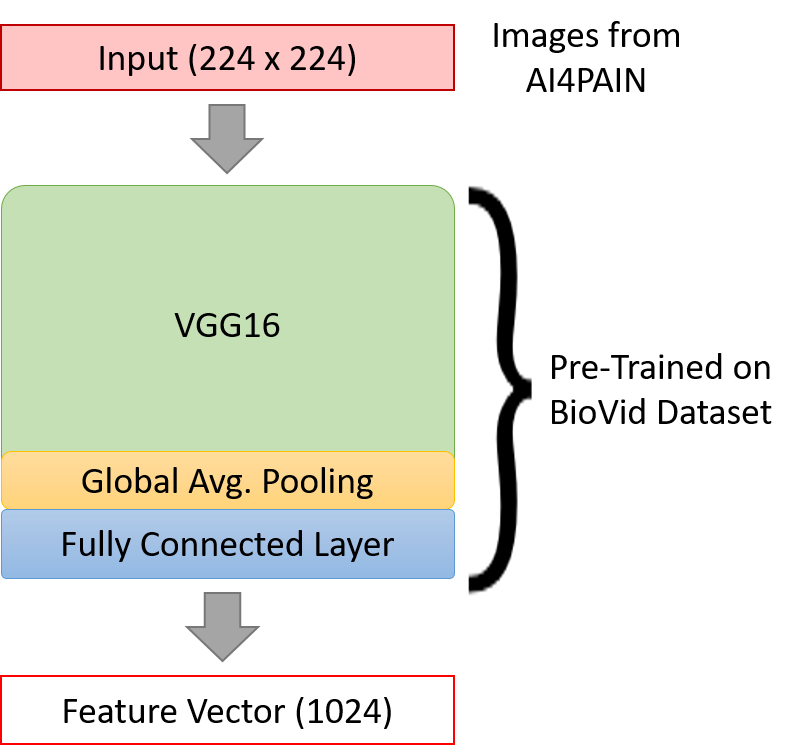}}
\caption{Visualization of our CNN feature extractor.}
\label{fig:feat_extract}
\end{figure}

The model follows a VGG16 architecture and takes as input face-cropped images scaled to 224 $\times$ 224 pixels. The final pooling layer of the original VGG16 architecture was modified to a Global Average Pooling instead of the Max Pooling. After the final pooling layer, the model is connected to a fully connected layer of size 1024, followed by the prediction layer. We utilize the model until the fully connected layer as the feature extractor (see Figure~\ref{fig:feat_extract}), i.e., the output of the fully connected layer serves as the feature vector (length $=$ 1024) for subsequent training. This model was trained on face images from the BioVid heat pain dataset~\cite{walter2013biovid}, which contains data from 87 participants. While the dataset contains five classes of pain, the model was trained as a binary classifier to differentiate between no-pain and high-pain images. The model achieved an accuracy of 0.70, and an F1-score of 0.69. We refer to the original model paper~\cite{prajod2022using} for the details of the training procedure and hyperparameters.

We note that the BioVid dataset has instances of heat pain, whereas the AI4Pain challenge dataset involves electrical pain. While these pain stimuli are different, the literature suggests that these stimuli lead to similar facial expressions~\cite{prkachin1992consistency, prkachin2009assessing, kunz2019facial}. Moreover, previous study by Othman~et~al.~\cite{othman2019cross} demonstrated that pain models trained on heat (BioVid) and electrical (X-ITE~\cite{gruss2019multi}) pain datasets have good cross-dataset performances. 

While we expect the BioVid model to be a good feature extractor, there are a couple of considerations that may affect the effective transferability of learned features. First, the chosen BioVid model was trained as a binary classifier (no-pain vs. high-pain). The subsequent performance and explainability assessments may have been influenced by this choice. In other words, while the BioVid model learned features to detect high-pain, it is not clear if these features are sufficient to differentiate low-pain and high-pain in the AI4Pain dataset. We address this gap by evaluating the performance of models trained on the AI4Pain dataset using the BioVid model as a feature extractor. Second, this model was trained and evaluated on images that were suggested by dataset creators as pain frames. Hence, the efficiency of this model in detecting the subtle changes in facial expressions during the entire pain experience is yet to be investigated. Similarly, the lack of time series information leads to the model not distinguishing between eye closures due to blinks and pain. Hence, our models are designed to capture the time series information for improved pain detection.

\subsection{Models}
\label{sec:training}

\subsubsection{Simple ANN with Majority Voting}
\begin{figure}[htbp]
\centerline{\includegraphics[width=0.7\linewidth]{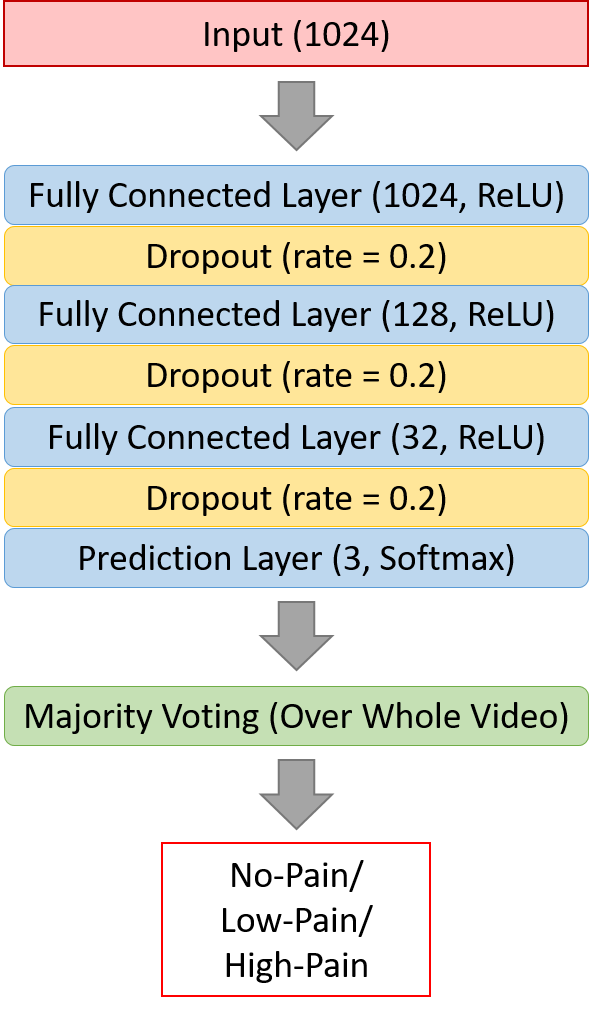}}
\caption{Visualization of our simple ANN model with majority voting scheme for video level prediction.}
\label{fig:ANN_arch}
\end{figure}

For our initial classifier, we use a straightforward neural network architecture to predict each frame in a video. The model comprises four fully connected layers with a decreasing number of neurons in each successive layer (1024, 128, 32, 3). Each layer employs the ReLU activation function. To mitigate overfitting, dropout is applied between each fully connected layer at a rate of 0.3. To generate a single label for the entire video, we predict each frame individually and then determine the final label through a majority vote. Figure~\ref{fig:ANN_arch} illustrates the model architecture for our simple neural network.

\subsubsection{LSTM}
\begin{figure}[htbp]
\centerline{\includegraphics[width=0.7\linewidth]{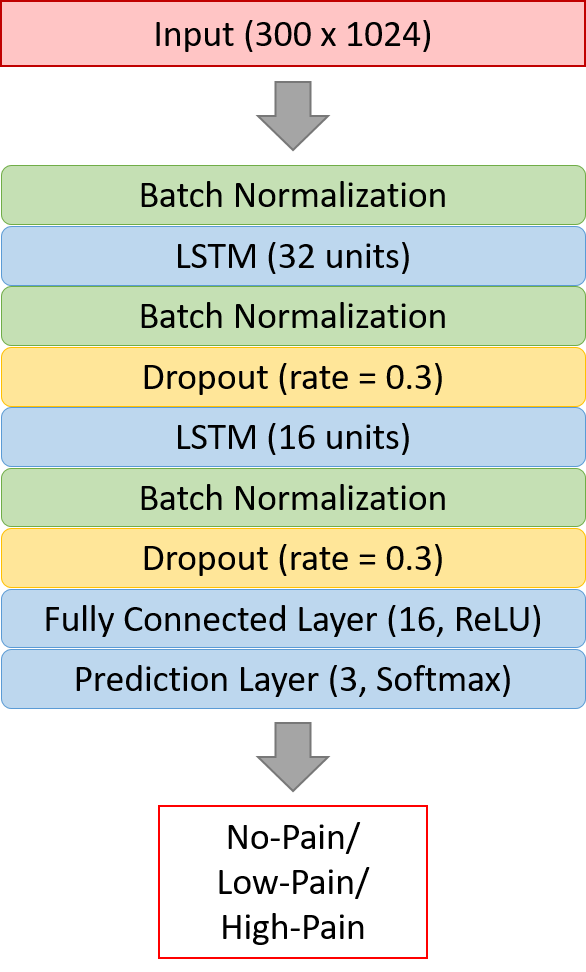}}
\caption{Visualization of our LSTM model for 10-second videos (300 frames).}
\label{fig:LSTM_arch}
\end{figure}

The LSTM layers in deep learning models are typically used to capture time series information. In this work, we utilize a relatively simple architecture involving two LSTM layers. The low-pain and high-pain videos were recorded at 30 frames per second and lasted for 10 seconds, resulting in 300 frames per video. So, the input to the model was 300 stacked arrays of length 1024 each, where each array represents features extracted for a frame of the video. A batch normalization layer was applied to the input to ensure stable training and faster convergence. The model contained two LSTM layers with 32 and 16 units. A batch normalization and a dropout layer (rate $=$ 0.3) were added after each LSTM layer to mitigate overfitting. Finally, the model was connected to a fully connected layer (16 units with ReLU activation) and a prediction layer (3 units with Softmax activation). The LSTM architecture is visualized in Figure~\ref{fig:LSTM_arch}.

The model was trained using weighted categorical cross-entropy loss and Adadelta (learning rate $=$ 1.0) optimizer. The loss function weights for each class were determined as inverse class frequencies. The training input was provided in batches of 32 and lasted for a maximum of 100 epochs. We employed the early stopping mechanism (patience $=$ 5) to avoid overfitting, i.e., the training stopped if the validation error did not reduce for five consecutive epochs.

Unlike low-pain and high-pain videos, the no-pain videos lasted around 60 seconds. The no-pain videos were divided into 10-second segments to maximize the training data, yielding 5-6 segments from the no-pain video. However, the validation and test sets utilized the first 300 frames of every video for comparing results with the baseline.

\section{Results and Discussion}

The challenge evaluates the performances using the Accuracy metric. We report the precision, recall, F1-score, and accuracy of the models on the validation set. Since the labels for the test set were not available, only the accuracy computed by the organizers is reported. To highlight our approach's effectiveness, we compare our model's performance on the validation and test sets to the baseline results on these sets.

\begin{table*}
\caption{Performance of our models and baseline on the validation set}
\centering
\label{tab:validation_perf}
\begin{tabular}{| c|c|c|c|c|c|c|c|c|c|c|c|c|c |} 
\hline
 \multirow{2}{6em}{Models}  & \multicolumn{4}{|c|}{Precision} & \multicolumn{4}{|c|}{Recall}  & \multicolumn{4}{|c|}{F1-score} & \multirow{2}{4em}{Accuracy}\\
\cline{2-13}
& No-Pain&  Low & High & Avg. & No-Pain & Low & High & Avg. & No-Pain & Low & High & Avg.& \\ 
\hline
Baseline (Video)  &  &   &&  &   &&  &  &  &   &&  & 0.40 \\ 
Baseline (fNIRS)  &  &   &&  &   &&  &  &  &   &&  & 0.43 \\ 
Baseline (Video + fNIRS)  &  &   &&  &   &&  &  &  &   &&  & 0.40 \\ 
Simple ANN + Voting & 0.10 & 0.60  & 0.66 & 0.45 & 0.17 & 0.67  &  0.55 & 0.46 & 0.12 & 0.63 & 0.60 & 0.45  & 0.59 con\\
LSTM & 0.24 & 0.59 & 0.71 & 0.51  & 0.42  & 0.74 & 0.49 & 0.55  & 0.30 & 0.65 & 0.58 & 0.51 & \textbf{0.60} \\
\hline
\end{tabular}
\end{table*}


Table~\ref{tab:validation_perf} presents the performance of models on the validation set. Interestingly, the fNIRS baseline model achieved the highest accuracy among all configurations of baseline evaluation. There was an decrease in the performance when video and fNIRS modalities were combined. Nevertheless, both our models outperformed all baseline models, with LSTM model achieving the highest accuracy of 0.60. Moreover, the class-wise recalls indicate that the chosen CNN features did not ignore one class, while achieving high prediction accuracy for other classes. In other words, the BioVid CNN features shows promising discerning capabilities for all three classes in the AI4Pain dataset.

\begin{table}
\caption{Performance of our models and baseline on the test set}
\centering
\label{tab:test_perf}
\begin{tabular}{| c|c |} 
\hline
Models  & Accuracy\\
\hline
Baseline (Video)  & 0.40 \\ 
Baseline (fNIRS)  & 0.43 \\ 
Baseline (Video + fNIRS)  & 0.42 \\ 
Simple ANN + Voting & \textbf{0.49} \\
LSTM & 0.43 \\
\hline
\end{tabular}
\end{table}

The performance of the models on the test set are presented in Table~\ref{tab:test_perf}. The baseline values are similar to that of the validation set. Again, the unimodal fNIRS model performed better than combining video and fNIRS. However, our models saw a considerable drop in accuracy, with around 0.1 drop in simple ANN accuracy and 0.17 drop in LSTM accuracy. Despite this drop, the LSTM model matched the best baseline performance, whereas the simple ANN model outperformed all the models.

We emphasized transferability through our approach and hence, adopted a unimodal pain recognition approach. However, typically multimodal data results in an improved recognition performance. Although the baseline performances show otherwise, the multimodal approach may potentially improve the performance with a different feature extraction or model architecture.

\section{Conclusion}

In this study, we explored the transferability of feature representation learned by a pain prediction model to another pain recognition model. Specifically, we assessed if the features learned by a heat pain model can be applied in training an electrical pain recognition model. Although the stimuli are different, the literature suggests that pain facial expressions caused by these stimuli share similarities. So, our assessment involved using an existing model trained on a heat pain dataset as a feature extractor. The facial images from the electrical pain dataset were fed to this feature extractor CNN to generate feature vector representations. These feature vectors were then used to train two electrical pain models: a simple ANN and an LSTM network. This assessment was carried out as a part of the AI4Pain challenge, which provided us the access to an electrical pain dataset. The challenge aimed at classifying participant data into three levels of pain: No-Pain, Low-Pain, and High-Pain. Although the challenge dataset includes videos and fNIRS data, we utilized only the video modality due to our focus on transferability. Our models outperformed the challenge baseline models on validation and test sets. Our results indicate that features learned during heat pain training can be leveraged to train electrical pain models. Our findings contribute to improving deep learned pain recognition with limited data, which is a typical characteristics of pain datasets. 




\bibliographystyle{IEEEtran}
\bibliography{main}

\begin{thebibliography}{10}
\providecommand{\url}[1]{#1}
\csname url@samestyle\endcsname
\providecommand{\newblock}{\relax}
\providecommand{\bibinfo}[2]{#2}
\providecommand{\BIBentrySTDinterwordspacing}{\spaceskip=0pt\relax}
\providecommand{\BIBentryALTinterwordstretchfactor}{4}
\providecommand{\BIBentryALTinterwordspacing}{\spaceskip=\fontdimen2\font plus
\BIBentryALTinterwordstretchfactor\fontdimen3\font minus \fontdimen4\font\relax}
\providecommand{\BIBforeignlanguage}[2]{{%
\expandafter\ifx\csname l@#1\endcsname\relax
\typeout{** WARNING: IEEEtran.bst: No hyphenation pattern has been}%
\typeout{** loaded for the language `#1'. Using the pattern for}%
\typeout{** the default language instead.}%
\else
\language=\csname l@#1\endcsname
\fi
#2}}
\providecommand{\BIBdecl}{\relax}
\BIBdecl

\bibitem{cowen2015assessing}
R.~Cowen, M.~K. Stasiowska, H.~Laycock, and C.~Bantel, ``Assessing pain objectively: the use of physiological markers,'' \emph{Anaesthesia}, vol.~70, no.~7, pp. 828--847, 2015.

\bibitem{raja2020revised}
S.~N. Raja, D.~B. Carr, M.~Cohen, N.~B. Finnerup, H.~Flor, S.~Gibson, F.~J. Keefe, J.~S. Mogil, M.~Ringkamp, K.~A. Sluka \emph{et~al.}, ``The revised international association for the study of pain definition of pain: concepts, challenges, and compromises,'' \emph{Pain}, vol. 161, no.~9, pp. 1976--1982, 2020.

\bibitem{cohen2018reconsidering}
M.~Cohen, J.~Quintner, and S.~Van~Rysewyk, ``Reconsidering the international association for the study of pain definition of pain,'' \emph{Pain reports}, vol.~3, no.~2, p. e634, 2018.

\bibitem{klingner2023mimik}
C.~M. Klingner and O.~Guntinas-Lichius, ``Mimik und emotion,'' \emph{Laryngo-rhino-otologie}, vol. 102, no. S 01, pp. S115--S125, 2023.

\bibitem{williams2011FacialEO}
A.~C. Williams, ``Facial expressions of pain: clinical meaning and research possibilities,'' 2011.

\bibitem{hadjistavropoulos2011biopsychosocial}
T.~Hadjistavropoulos, K.~D. Craig, S.~Duck, A.~Cano, L.~Goubert, P.~L. Jackson, J.~S. Mogil, P.~Rainville, M.~J. Sullivan, A.~C. d.~C. Williams \emph{et~al.}, ``A biopsychosocial formulation of pain communication.'' \emph{Psychological bulletin}, vol. 137, no.~6, p. 910, 2011.

\bibitem{zamzmi2018neonatal}
G.~Zamzmi, D.~Goldgof, R.~Kasturi, and Y.~Sun, ``Neonatal pain expression recognition using transfer learning,'' \emph{arXiv preprint arXiv:1807.01631}, 2018.

\bibitem{egede2017fusing}
J.~Egede, M.~Valstar, and B.~Martinez, ``Fusing deep learned and hand-crafted features of appearance, shape, and dynamics for automatic pain estimation,'' in \emph{12th IEEE international conference on automatic face \& gesture recognition (FG 2017)}.\hskip 1em plus 0.5em minus 0.4em\relax IEEE, 2017, pp. 689--696.

\bibitem{wang2017regularizing}
F.~Wang, X.~Xiang, C.~Liu, T.~D. Tran, A.~Reiter, G.~D. Hager, H.~Quon, J.~Cheng, and A.~L. Yuille, ``Regularizing face verification nets for pain intensity regression,'' in \emph{2017 IEEE International Conference on Image Processing (ICIP)}.\hskip 1em plus 0.5em minus 0.4em\relax IEEE, 2017, pp. 1087--1091.

\bibitem{hassan2019automatic}
T.~Hassan, D.~Seu{\ss}, J.~Wollenberg, K.~Weitz, M.~Kunz, S.~Lautenbacher, J.-U. Garbas, and U.~Schmid, ``Automatic detection of pain from facial expressions: a survey,'' \emph{IEEE transactions on pattern analysis and machine intelligence}, vol.~43, no.~6, pp. 1815--1831, 2019.

\bibitem{kunz2017problems}
M.~Kunz, D.~Seuss, T.~Hassan, J.~U. Garbas, M.~Siebers, U.~Schmid, M.~Sch{\"o}berl, and S.~Lautenbacher, ``Problems of video-based pain detection in patients with dementia: a road map to an interdisciplinary solution,'' \emph{BMC geriatrics}, vol.~17, no.~1, pp. 1--8, 2017.

\bibitem{prajod2022deep}
P.~Prajod, D.~Schiller, T.~Huber, and E.~Andr{\'e}, ``Do deep neural networks forget facial action units?—exploring the effects of transfer learning in health related facial expression recognition,'' \emph{AI for Disease Surveillance and Pandemic Intelligence: Intelligent Disease Detection in Action}, vol. 1013, p. 217, 2022.

\bibitem{xiang2022imbalanced}
X.~Xiang, F.~Wang, Y.~Tan, and A.~L. Yuille, ``Imbalanced regression for intensity series of pain expression from videos by regularizing spatio-temporal face nets,'' \emph{Pattern Recognition Letters}, vol. 163, pp. 152--158, 2022.

\bibitem{challenge_paper}
R.~Fernandez~Rojas, N.~Hirachan, C.~Joseph, B.~Seymour, and R.~Goecke, ``The ai4pain grand challenge 2024: Advancing pain assessment with multmodal fnirs and facial video analysis,'' in \emph{2024 12th Internatonal Conference on Afectve Computng and Intelligent Interacton}.\hskip 1em plus 0.5em minus 0.4em\relax IEEE, 2024.

\bibitem{valstar201711}
M.~Valstar, S.~Zafeiriou, and M.~Pantic, ``Facial actions as social signals,'' \emph{Social signal processing}, p. 123, 2017.

\bibitem{ko2018brief}
B.~C. Ko, ``A brief review of facial emotion recognition based on visual information,'' \emph{sensors}, vol.~18, no.~2, p. 401, 2018.

\bibitem{prkachin2009assessing}
K.~M. Prkachin \emph{et~al.}, ``Assessing pain by facial expression: facial expression as nexus,'' \emph{Pain Research and Management}, vol.~14, pp. 53--58, 2009.

\bibitem{florea2014learning}
C.~Florea, L.~Florea, and C.~Vertan, ``Learning pain from emotion: transferred hot data representation for pain intensity estimation,'' in \emph{European Conference on Computer Vision}.\hskip 1em plus 0.5em minus 0.4em\relax Springer, 2014, pp. 778--790.

\bibitem{gkikas2023automatic}
S.~Gkikas and M.~Tsiknakis, ``Automatic assessment of pain based on deep learning methods: A systematic review,'' \emph{Computer methods and programs in biomedicine}, vol. 231, p. 107365, 2023.

\bibitem{prajod2022using}
P.~Prajod, T.~Huber, and E.~Andr{\'e}, ``Using explainable ai to identify differences between clinical and experimental pain detection models based on facial expressions,'' in \emph{International Conference on Multimedia Modeling}.\hskip 1em plus 0.5em minus 0.4em\relax Springer, 2022, pp. 311--322.

\bibitem{rodriguez2017deep}
P.~Rodriguez, G.~Cucurull, J.~Gonz{\`a}lez, J.~M. Gonfaus, K.~Nasrollahi, T.~B. Moeslund, and F.~X. Roca, ``Deep pain: Exploiting long short-term memory networks for facial expression classification,'' \emph{IEEE transactions on cybernetics}, 2017.

\bibitem{haque2018deep}
M.~A. Haque, R.~B. Bautista, F.~Noroozi, K.~Kulkarni, C.~B. Laursen, R.~Irani, M.~Bellantonio, S.~Escalera, G.~Anbarjafari, K.~Nasrollahi \emph{et~al.}, ``Deep multimodal pain recognition: a database and comparison of spatio-temporal visual modalities,'' in \emph{13th IEEE International Conference on Automatic Face \& Gesture Recognition (FG 2018)}.\hskip 1em plus 0.5em minus 0.4em\relax IEEE, 2018, pp. 250--257.

\bibitem{kunz2019facial}
M.~Kunz, D.~Meixner, and S.~Lautenbacher, ``Facial muscle movements encoding pain—a systematic review,'' \emph{Pain}, vol. 160, no.~3, pp. 535--549, 2019.

\bibitem{othman2019cross}
E.~Othman, P.~Werner, F.~Saxen, A.~Al-Hamadi, and S.~Walter, ``Cross-database evaluation of pain recognition from facial video,'' in \emph{2019 11th International Symposium on Image and Signal Processing and Analysis (ISPA)}.\hskip 1em plus 0.5em minus 0.4em\relax IEEE, 2019, pp. 181--186.

\bibitem{dai2019real}
L.~Dai, J.~Broekens, and K.~P. Truong, ``Real-time pain detection in facial expressions for health robotics,'' in \emph{2019 8th International Conference on Affective Computing and Intelligent Interaction Workshops and Demos (ACIIW)}.\hskip 1em plus 0.5em minus 0.4em\relax IEEE, 2019, pp. 277--283.

\bibitem{tavakolian2020self}
M.~Tavakolian, M.~B. Lopez, and L.~Liu, ``Self-supervised pain intensity estimation from facial videos via statistical spatiotemporal distillation,'' \emph{Pattern Recognition Letters}, vol. 140, pp. 26--33, 2020.

\bibitem{fernandez2023multimodal}
R.~Fernandez~Rojas, N.~Hirachan, N.~Brown, G.~Waddington, L.~Murtagh, B.~Seymour, and R.~Goecke, ``Multimodal physiological sensing for the assessment of acute pain,'' \emph{Frontiers in Pain Research}, vol.~4, p. 1150264, 2023.

\bibitem{face-alignment}
A.~Bulat and G.~Tzimiropoulos, ``How far are we from solving the 2d \& 3d face alignment problem? (and a dataset of 230,000 3d facial landmarks),'' in \emph{International Conference on Computer Vision}, 2017.

\bibitem{coutrin2022convolutional}
G.~A. Coutrin, L.~P. Carlini, L.~A. Ferreira, T.~M. Heiderich, R.~C. Balda, M.~C. Barros, R.~Guinsburg, and C.~E. Thomaz, ``Convolutional neural networks for newborn pain assessment using face images: A quantitative and qualitative comparison,'' in \emph{International Conference on Medical Imaging and Computer-Aided Diagnosis}.\hskip 1em plus 0.5em minus 0.4em\relax Springer, 2022, pp. 503--513.

\bibitem{walter2013biovid}
S.~Walter, S.~Gruss, H.~Ehleiter, J.~Tan, H.~C. Traue, P.~Werner, A.~Al-Hamadi, S.~Crawcour, A.~O. Andrade, and G.~M. da~Silva, ``The biovid heat pain database data for the advancement and systematic validation of an automated pain recognition system,'' in \emph{2013 IEEE international conference on cybernetics (CYBCO)}.\hskip 1em plus 0.5em minus 0.4em\relax IEEE, 2013, pp. 128--131.

\bibitem{prkachin1992consistency}
K.~M. Prkachin, ``The consistency of facial expressions of pain: a comparison across modalities,'' \emph{Pain}, vol.~51, no.~3, pp. 297--306, 1992.

\bibitem{gruss2019multi}
S.~Gruss, M.~Geiger, P.~Werner, O.~Wilhelm, H.~C. Traue, A.~Al-Hamadi, and S.~Walter, ``Multi-modal signals for analyzing pain responses to thermal and electrical stimuli,'' \emph{JoVE (Journal of Visualized Experiments)}, no. 146, p. e59057, 2019.

\end{thebibliography}

\end{document}